\documentclass{article}

     \PassOptionsToPackage{numbers, compress}{natbib}
     \usepackage[final]{neurips_2019}

\usepackage[utf8]{inputenc} % allow utf-8 input
\usepackage[T1]{fontenc}    % use 8-bit T1 fonts
\usepackage{hyperref}       % hyperlinks
\usepackage{url}            % simple URL typesetting
\usepackage{booktabs}       % professional-quality tables
\usepackage{amsfonts}       % blackboard math symbols
\usepackage{nicefrac}       % compact symbols for 1/2, etc.
\usepackage{microtype}      % microtypography

\usepackage{amsmath}
\usepackage{tikz}
\usepackage{subcaption}
\usepackage{placeins}

\title{Make Thunderbolts Less Frightening --- Predicting Extreme Weather Using Deep Learning}

\author{%
	Christian~Sch\"on \\
	Big Data Analytics Group \\
	Saarland Informatics Campus\\
	bigdata.uni-saarland.de \\
	\And
	Jens Dittrich \\
	Big Data Analytics Group \\
	Saarland Informatics Campus \\
	bigdata.uni-saarland.de \\
}

\begin{document}

\maketitle

\begin{abstract}
  Forecasting severe weather conditions is still a very challenging and computationally expensive task due to the enormous amount of data and the complexity of the underlying physics. Machine learning approaches and especially deep learning have however shown huge improvements in many research areas dealing with large datasets in recent years. In this work, we tackle one specific sub-problem of weather forecasting, namely the prediction of thunderstorms and lightning. We propose the use of a convolutional neural network architecture inspired by UNet++ and ResNet to predict thunderstorms as a binary classification problem based on satellite images and lightnings recorded in the past. We achieve a probability of detection of more than 94\% for lightnings within the next 15 minutes while at the same time minimizing the false alarm ratio compared to previous approaches.
\end{abstract}

\section{Introduction}

Climate change will affect humanity in many ways over the next decades. Besides well known effects such as the melting polar caps and the increasing sea levels, recent studies also predict an increasing number of convective storms over the United States~\cite{seeley2015effect,hoogewind2017impact,diffenbaugh2013robust} and Europe~\cite{pucik2017future}. Convective storms are often accompanied by lightning, heavy rain, hail, and sometimes also tornadoes. In fields such as aviation, thunderstorms can pose a real security threat if planes are not warned and detoured in time. Predicting such severe weather conditions is therefore a core task for weather services.
However, even state-of-the-art systems such as NowCastMIX~\cite{james2018nowcastmix}, a system operated by the German Meteorological Service, still struggle with a high false alarm ratio, especially if the forecast period is increased to one hour and beyond.

Most weather models currently in operational mode are based on numerical weather prediction~(NWP) and estimate the state of the atmosphere by applying transformations based on physical laws. Machine learning algorithms have however shown recent success in many fields, especially those providing large datasets for training. The availability of geostationary satellites, lightning detection networks and other data sources therefore encourages researchers to investigate the application of machine learning in the context of severe weather prediction. 

In this work, we are focusing on one small sub-problem, namely the prediction of lightning and thunderstorms. We propose the use of a convolutional neural network architecture inspired by UNet++~\cite{zhou2018unet++} and ResNet~\cite{he2016deep}, very similar to the architecture used by Peng et al.~\cite{peng2019end}.
Based on satellite images taken by the SEVIRI instrument onboard the current, second generation of Meteosat satellites~\cite{schmetz2002introduction} and past lightning observations registered by the LINET network~\cite{Betz2009Linet}, we try to predict whether there will be lightning or not in the future.

\section{Related Work}

Our approach is mainly based on satellite images, making convolutional neural networks a natural choice for the network. Considering the fact that we try to distinguish areas affected by thunderstorms and areas without such events, our work is closely related to the idea of image segmentation. Among the many architectures proposed in this field throughout the last years, encoder-decoder networks have become state-of-the-art. Fully convolutional networks (FCN)~\cite{long2015fully} and UNet~\cite{ronneberger2015unet} are among the most successful and influential ones introducing the idea of skip connections. Follow-up works extended these base networks in different ways, among them H-DenseUNet~\cite{li2018h} and UNet++~\cite{zhou2018unet++} which are inspired by dense connections between subsequent layers introduced in DenseNet~\cite{gao2016dense}.
In the related field of image classification, ResNet~\cite{he2016deep} introduced the idea of residual learning. Peng et al.~\cite{peng2019end} adopted the idea of residual learning for image segmentation tasks, introducing residual blocks in the UNet++ architecture. Our architecture follows the same idea and differs only in details.

In the field of convection and thunderstorm prediction, several research groups have published first results using machine learning approaches. Some of these approaches are based on Random Forests as introduced by Breiman~\cite{breiman2001random}. Ahijevych et al.~\cite{williams2016probabilistic} used parameters of a convection permitting model to predict convective system initiation. More recent works focus directly on the prediction of lightings. In a previous work~\cite{schoen2019error}, we proposed to use the error resulting from satellite image nowcasting as a feature for lightning prediction based on Random Forests. Geng et al.~\cite{geng2019lightnet} used a different approach relying on NWP model parameters, lightning observations and a recurrent convolutional neural network.

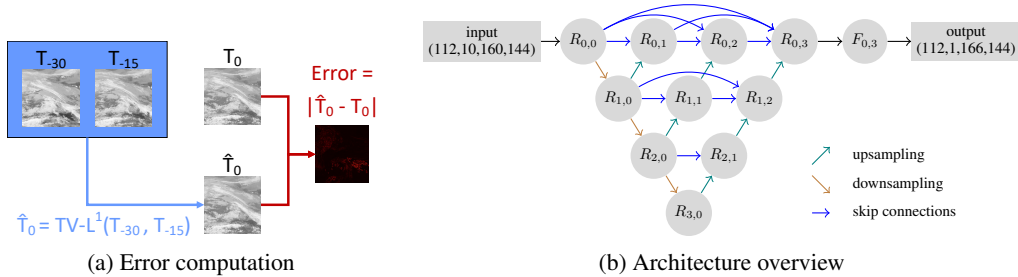
\begin{figure}
	\centering
	\begin{subfigure}[t]{0.35\textwidth}
		\centering
		\includegraphics[trim=5.0mm 5.0mm 5.0mm 5.0mm, clip, width=\linewidth, keepaspectratio]{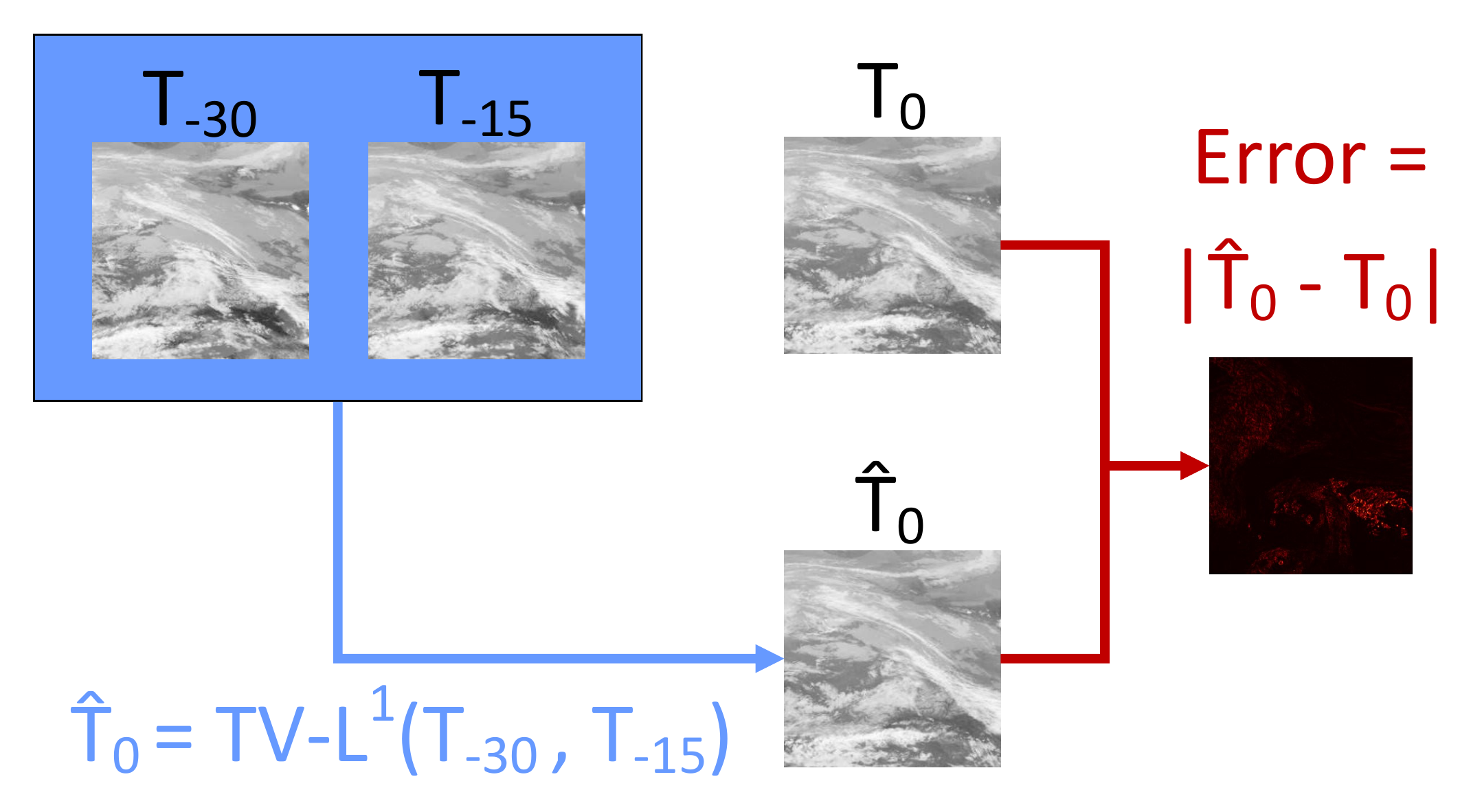}
		\subcaption[Error Computation]{Error computation}
		\label{subfig:Data_Preprocessing}
	\end{subfigure}
	\hfill
	\begin{subfigure}[t]{0.64\textwidth}
		\centering
		\begin{tikzpicture}[scale=0.585, transform shape]
		\tikzstyle{basicblock} = [circle, fill=gray!30]
		\tikzstyle{io} = [rectangle, fill=gray!30, align=center]
		
		\node[io] (i) at (-0.3, 3.9) {input\\(112,10,160,144)};
		
		\node[basicblock] (x00) at (2.0, 3.9) {$R_{0,0}$};
		\node[basicblock] (x01) at (3.6, 3.9) {$R_{0,1}$};
		\node[basicblock] (x02) at (5.2, 3.9) {$R_{0,2}$};
		\node[basicblock] (x03) at (6.8, 3.9) {$R_{0,3}$};
		\node[basicblock] (x10) at (2.8, 2.6) {$R_{1,0}$};
		\node[basicblock] (x11) at (4.4, 2.6) {$R_{1,1}$};
		\node[basicblock] (x12) at (6, 2.6) {$R_{1,2}$};
		\node[basicblock] (x20) at (3.6, 1.3) {$R_{2,0}$};
		\node[basicblock] (x21) at (5.2, 1.3) {$R_{2,1}$};
		\node[basicblock] (x30) at (4.4, 0) {$R_{3,0}$};
		
		\node[basicblock] (f03) at (8.4, 3.9) {$F_{0,3}$};
		
		\node[io] (o) at (10.7, 3.9) {output\\(112,1,166,144)};
		
		\foreach \from/\to in {x00/x10, x10/x20, x20/x30}
		\draw [->, brown] (\from) -- (\to);
		
		\foreach \from/\to in {x30/x21, x20/x11, x21/x12, x10/x01, x11/x02, x12/x03}
		\draw [->, teal] (\from) -- (\to);
		
		\foreach \from/\to in {x00/x01, x01/x02, x02/x03, x10/x11, x11/x12, x20/x21}
		\draw [->, blue] (\from) -- (\to);
		
		\foreach \from/\to in {x00/x02, x00/x03, x01/x03, x10/x12}
		\draw [->, blue] (\from) to[bend left] (\to);
		
		\draw [->] (i) -- (x00);
		\draw [->] (f03) -- (o);
		
		\draw [->] (x03) -- (f03);
		
		\draw [->, blue] (7.2, 0) -- (7.6, 0);
		\node[draw=none, fill=none, right] (l3) at (8.0, 0) {skip connections};
		\draw [->, brown] (7.2, 0.85) -- (7.6, 0.45);
		\node[draw=none, fill=none, right] (l1) at (8.0, 0.65) {downsampling};
		\draw [->, teal] (7.2, 1.1) -- (7.6, 1.5);
		\node[draw=none, fill=none, right] (l2) at (8.0, 1.3) {upsampling};
		\end{tikzpicture}
		\subcaption[Architecture Overview]{Architecture overview}
		\label{subfig:Architecture_Overview}
	\end{subfigure}
	\caption{Data preprocessing \& network architecture}
\end{figure}

\section{Dataset \& Preprocessing}

We follow the idea presented in our previous work~\cite{schoen2019error}, i.e.~using the error of satellite image nowcasting as a feature for lightning prediction. Our preprocessing pipeline depicted in \autoref{subfig:Data_Preprocessing} is therefore comparable: The error is computed by applying the optical flow algorithm TV-L$^1$~\cite{zach2007duality} to two consecutive satellite images T$_{\text{-30}}$ and T$_{\text{-15}}$ and taking the absolute difference to the original image at~T$_{\text{0}}$. The lightning data are accumulated in maps with the same temporal and spatial resolution. In contrast to their work, we intend to use images as input to our network and therefore omit the steps of splitting the data into single tiles and applying manual convolution. We also add the last lightning observations as an additional feature which has not been considered by Sch\"on et al., leading to a total of ten feature channels which we normalize to the range of [0,1].

Our dataset consists of satellite images and lightning observations taken between 2017-06-01 and 2017-07-04. Each original image of size $1114 \times 956$ is split into 56 samples of size $160 \times 144$. As we have conducted a cross validation experiment, we split the complete data range into four test sets, each surrounded by a twelve hour margin to avoid cross correlation effects with the training data. The resulting test sets and their sizes can be seen in \autoref{tab:Cross_Validation_Sets} in the Appendix.

\section{Network Architecture \& Training}

Our network architecture closely follows the proposal of Peng et al.~\cite{peng2019end} by combining UNet++~\cite{zhou2018unet++} with residual building blocks. An overview of the architecture for inference is depicted in \autoref{subfig:Architecture_Overview}. Each node $R$ in the figure is a single residual block. If a node has more than one incoming edge, the tensors are simply concatenated. The details of the architecture can be found in \autoref{app:Network_Architecture_Details}.
To tackle the problem of imbalanced classes, we use a weighted binary cross entropy loss and set the weight for the positive class to the ratio \textit{\# negative samples : \# positive samples}.
Details about our architecture and experimental setup can be found in Appendix \ref{app:Network_Architecture_Details} and \ref{app:Training_Settings}.

\section{Results}

\begin{figure}
	\centering
	\begin{subfigure}[t]{0.65\textwidth}
		\centering
		\includegraphics[trim=3.8mm 9.0mm 15.8mm 9.3mm, clip, width=\linewidth, keepaspectratio]{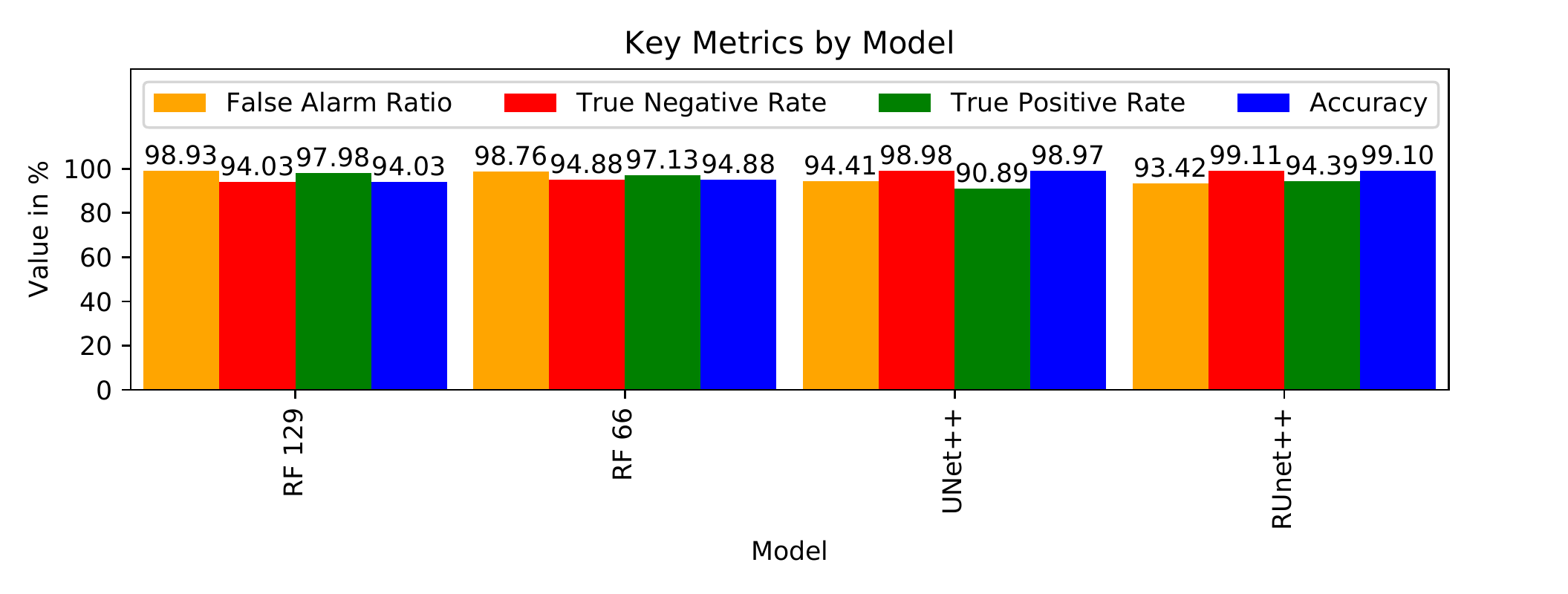}
		\subcaption[Key Metrics]{False Alarm Ratio, True Negative Rate, True Positive Rate and Accuracy per model}
		\label{fig:Key_Metrics}
	\end{subfigure}
	\hfill
	\begin{subfigure}[t]{0.325\textwidth}
		\centering
		\includegraphics[trim=3.8mm 3.0mm 4.6mm 9.3mm, clip, width=\linewidth, keepaspectratio]{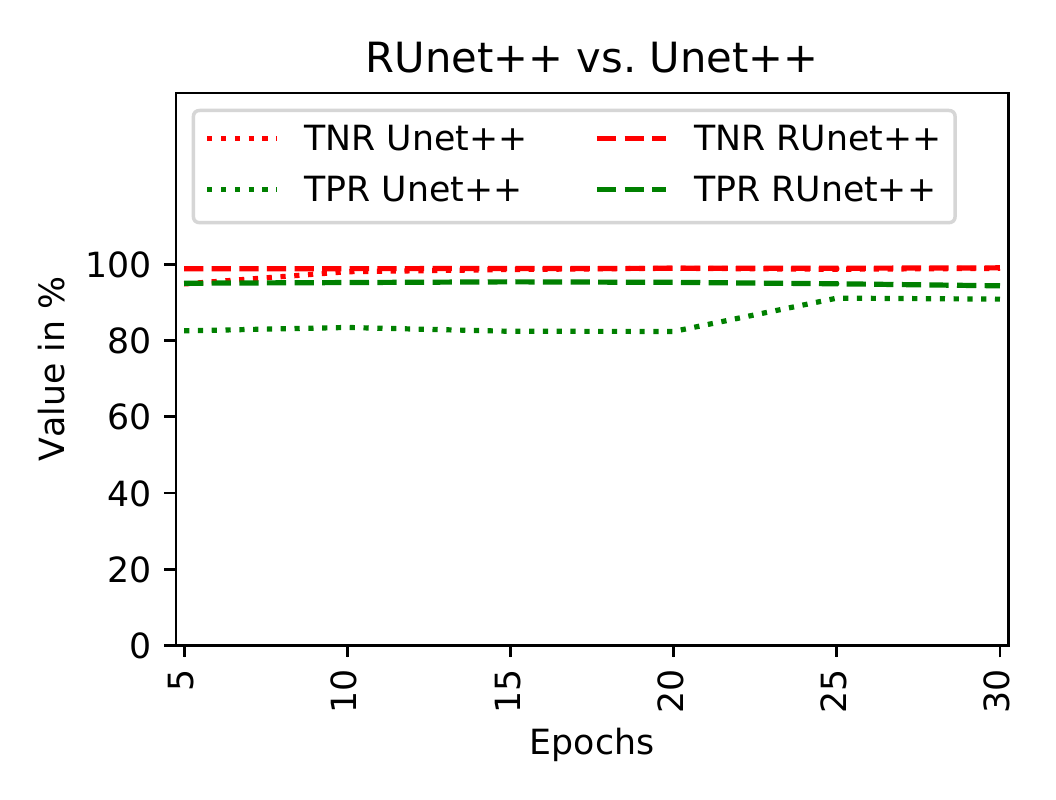}
		\subcaption[Key Metrics]{True Positive Rate and True Negative Rate per epoch}
		\label{fig:TPR_TNR_Training}
	\end{subfigure}
	\caption{Comparison of True Positive Rate, True Negative Rate, Accuracy, and False Alarm Ratio}
\end{figure}

The main metrics for our evaluation are True Positive Rate (TPR), True Negative Rate (TNR), Accuracy and False Alarm Ratio (FAR) whose formulas are given in \autoref{app:Evaluation_Metrics}. \autoref{fig:Key_Metrics} depicts the accumulated results over all four cross validation test sets. The RUNet++ bars correspond to our newly tested architecture whereas RF 129 and RF 66 are Random Forests as described in our previous work~\cite{schoen2019error}. The Random Forests were trained on a balanced subset of the data. 
As the ratio of positive samples is 0.066\%, the dataset has approximately 1,500 negative samples per positive sample and we can recompute the accuracy and FAR over all data by re-weighting the TNR with this factor.
To compare the impact of residual blocks in the UNet++ architecture, we also trained a standard UNet++ architecture with the same size as our new, residual version. We eliminated the first $1 \times 1$ convolution layer (i.e.~a layer that simply scales the input linearly) as well as the skip connection, leaving a basic building block consisting of two convolutions, normalizations and activations each.

Our results indicate that UNet++ and RUNet++ in general achieve a similar performance compared to Random Forests. However, in a direct comparison, they tend to predict the negative class better, resulting in a lower FAR which highly depends on the number of false positives in such an extremely imbalanced dataset. Reducing the FAR without sacrificing too much TPR is essential for the operational use as humans tend to mistrust systems that often fail. In addition, convolutional neural networks offer the advantage of directly processing image slices instead of single pixel values as compared to Random Forests, eliminating the need of additional preprocessing steps.

Comparing the standard UNet++ architecture with our modified version RUNet++, we can see that our model outperforms the standard architecture by 3.5\% in terms of TPR. We evaluated the test set every 5 epochs, which allows us to have a closer look at the development of the TPR and TNR during the model training. \autoref{fig:TPR_TNR_Training} indicates that although both models achieve a similar overall performance, RUNet++ converges much faster, which allows for a faster training.

\section{Conclusion}

Compared to previous works in the domain of lightning prediction, our approach shows promising results. It beats the results by Sch\"on et al. in terms of accuracy and false alarm ratio. A comparison with the work of Geng et al. is more difficult as the underlying dataset is different, but our approach shows that similar results can be achieved without the use of computationally expensive NWP model parameters.
The results presented in this paper are only a first investigation of convolutional neural networks in the context of lightning prediction. There is still future work to do, such as extending the forecast period and adding new features and data to confirm the capability of the system to generalize to other years and seasons.
The high false alarm ratio might be minimized by putting more weight to the negative class during loss computation, potentially leading to fewer false positives at the cost of a worse detection of lightnings.

\subsubsection*{Acknowledgments}

All work was done as part of the project FE-Nr. 50.0383/2018 for the German Federal Ministry of Transport and Digital Infrastructure. The authors are solely responsible for the contents.

%\newpage
\small{
\bibliographystyle{acm}
\bibliography{bibliography}

\begin{thebibliography}{10}

\bibitem{williams2016probabilistic}
{\sc Ahijevych, D., Pinto, J.~O., Williams, J.~K., and Steiner, M.}
\newblock Probabilistic forecasts of mesoscale convective system initiation
  using the random forest data mining technique.
\newblock {\em Weather and Forecasting 31}, 2 (2016), 581--599.

\bibitem{Betz2009Linet}
{\sc Betz, H.~D., Schmidt, K., Laroche, P., Blanchet, P., Oettinger, W.~P.,
  Defer, E., Dziewit, Z., and Konarski, J.}
\newblock Linet — an international lightning detection network in europe.
\newblock {\em Atmospheric Research 91}, 2 (2009), 564 -- 573.

\bibitem{breiman2001random}
{\sc Breiman, L.}
\newblock Random forests.
\newblock {\em Machine learning 45}, 1 (2001), 5--32.

\bibitem{diffenbaugh2013robust}
{\sc Diffenbaugh, N.~S., Scherer, M., and Trapp, R.~J.}
\newblock Robust increases in severe thunderstorm environments in response to
  greenhouse forcing.
\newblock {\em Proceedings of the National Academy of Sciences 110}, 41 (2013),
  16361--16366.

\bibitem{geng2019lightnet}
{\sc Geng, Y.-a., Li, Q., Lin, T., Jiang, L., Xu, L., Zheng, D., Yao, W., Lyu,
  W., and Zhang, Y.}
\newblock Lightnet: A dual spatiotemporal encoder network model for lightning
  prediction.
\newblock In {\em Proceedings of the 25th ACM SIGKDD International Conference
  on Knowledge Discovery \& Data Mining\/} (2019), ACM, pp.~2439--2447.

\bibitem{he2016deep}
{\sc He, K., Zhang, X., Ren, S., and Sun, J.}
\newblock Deep residual learning for image recognition.
\newblock In {\em Proceedings of the IEEE conference on computer vision and
  pattern recognition\/} (2016), pp.~770--778.

\bibitem{hoogewind2017impact}
{\sc Hoogewind, K.~A., Baldwin, M.~E., and Trapp, R.~J.}
\newblock The impact of climate change on hazardous convective weather in the
  united states: Insight from high-resolution dynamical downscaling.
\newblock {\em Journal of Climate 30}, 24 (2017), 10081--10100.

\bibitem{gao2016dense}
{\sc Huang, G., Liu, Z., and Weinberger, K.~Q.}
\newblock Densely connected convolutional networks.
\newblock {\em CoRR abs/1608.06993\/} (2016).

\bibitem{james2018nowcastmix}
{\sc James, P.~M., Reichert, B.~K., and Heizenreder, D.}
\newblock Nowcastmix: Automatic integrated warnings for severe convection on
  nowcasting time scales at the german weather service.
\newblock {\em Weather and Forecasting 33}, 5 (2018), 1413--1433.

\bibitem{li2018h}
{\sc Li, X., Chen, H., Qi, X., Dou, Q., Fu, C.-W., and Heng, P.-A.}
\newblock H-denseunet: hybrid densely connected unet for liver and tumor
  segmentation from ct volumes.
\newblock {\em IEEE transactions on medical imaging 37}, 12 (2018), 2663--2674.

\bibitem{long2015fully}
{\sc Long, J., Shelhamer, E., and Darrell, T.}
\newblock Fully convolutional networks for semantic segmentation.
\newblock In {\em Proceedings of the IEEE conference on computer vision and
  pattern recognition\/} (2015), pp.~3431--3440.

\bibitem{peng2019end}
{\sc Peng, D., Zhang, Y., and Guan, H.}
\newblock End-to-end change detection for high resolution satellite images
  using improved unet++.
\newblock {\em Remote Sensing 11}, 11 (Jun 2019), 1382.

\bibitem{pucik2017future}
{\sc Púčik, T., Groenemeijer, P., Rädler, A.~T., Tijssen, L., Nikulin, G.,
  Prein, A.~F., van Meijgaard, E., Fealy, R., Jacob, D., and Teichmann, C.}
\newblock Future changes in european severe convection environments in a
  regional climate model ensemble.
\newblock {\em Journal of Climate 30}, 17 (2017), 6771--6794.

\bibitem{ronneberger2015unet}
{\sc Ronneberger, O., Fischer, P., and Brox, T.}
\newblock U-net: Convolutional networks for biomedical image segmentation.
\newblock In {\em Medical Image Computing and Computer-Assisted Intervention -
  {MICCAI} 2015 - 18th International Conference Munich, Germany, October 5 - 9,
  2015, Proceedings, Part {III}\/} (2015), pp.~234--241.

\bibitem{schmetz2002introduction}
{\sc Schmetz, J., Pili, P., Tjemkes, S., Just, D., Kerkmann, J., Rota, S., and
  Ratier, A.}
\newblock An introduction to meteosat second generation (msg).
\newblock {\em Bulletin of the American Meteorological Society 83}, 7 (2002),
  977--992.

\bibitem{schoen2019error}
{\sc Sch{\"{o}}n, C., Dittrich, J., and M{\"{u}}ller, R.}
\newblock The error is the feature: How to forecast lightning using a model
  prediction error.
\newblock In {\em Proceedings of the 25th {ACM} {SIGKDD} International
  Conference on Knowledge Discovery {\&} Data Mining\/} (2019), ACM,
  pp.~2979--2988.

\bibitem{seeley2015effect}
{\sc Seeley, J.~T., and Romps, D.~M.}
\newblock The effect of global warming on severe thunderstorms in the united
  states.
\newblock {\em Journal of Climate 28}, 6 (2015), 2443--2458.

\bibitem{zach2007duality}
{\sc Zach, C., Pock, T., and Bischof, H.}
\newblock A duality based approach for realtime {TV-L$^1$} optical flow.
\newblock In {\em Proceedings of the 29th DAGM Conference on Pattern
  Recognition\/} (2007), Springer-Verlag, pp.~214--223.

\bibitem{zhou2018unet++}
{\sc Zhou, Z., Siddiquee, M. M.~R., Tajbakhsh, N., and Liang, J.}
\newblock Unet++: A nested u-net architecture for medical image segmentation.
\newblock In {\em Deep Learning in Medical Image Analysis and Multimodal
  Learning for Clinical Decision Support}. Springer, 2018, pp.~3--11.

\end{thebibliography}
}

\FloatBarrier
%\newpage
\appendix

\section{Network Architecture Details}
\label{app:Network_Architecture_Details}

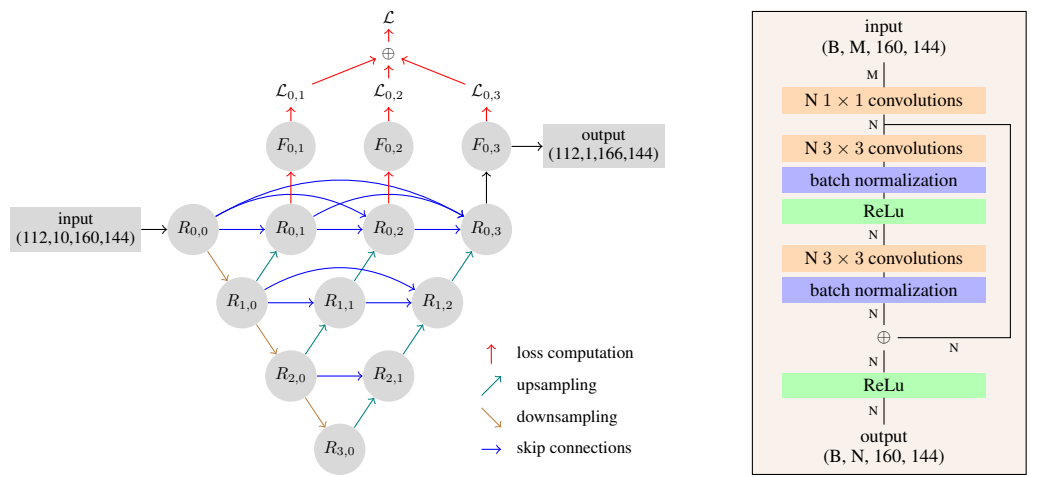
\begin{figure}
	\centering
	\begin{subfigure}[t]{0.63\textwidth}
		\centering
		\begin{tikzpicture}[scale=0.65, transform shape]
			\tikzstyle{basicblock} = [circle, fill=gray!30]
			\tikzstyle{io} = [rectangle, fill=gray!30, align=center]
			
			\node[io] (i) at (0, 4.5) {input\\(112,10,160,144)};
			
			\node[basicblock] (x00) at (2.4, 4.5) {$R_{0,0}$};
			\node[basicblock] (x01) at (4.4, 4.5) {$R_{0,1}$};
			\node[basicblock] (x02) at (6.4, 4.5) {$R_{0,2}$};
			\node[basicblock] (x03) at (8.4, 4.5) {$R_{0,3}$};
			\node[basicblock] (x10) at (3.4, 3) {$R_{1,0}$};
			\node[basicblock] (x11) at (5.4, 3) {$R_{1,1}$};
			\node[basicblock] (x12) at (7.4, 3) {$R_{1,2}$};
			\node[basicblock] (x20) at (4.4, 1.5) {$R_{2,0}$};
			\node[basicblock] (x21) at (6.4, 1.5) {$R_{2,1}$};
			\node[basicblock] (x30) at (5.4, 0) {$R_{3,0}$};
			
			\node[basicblock] (f01) at (4.4, 6.2) {$F_{0,1}$};
			\node[basicblock] (f02) at (6.4, 6.2) {$F_{0,2}$};
			\node[basicblock] (f03) at (8.4, 6.2) {$F_{0,3}$};
			
			\node (l01) at (4.4, 7.3) {$\mathcal{L}_{0,1}$};
			\node (l02) at (6.4, 7.3) {$\mathcal{L}_{0,2}$};
			\node (l03) at (8.4, 7.3) {$\mathcal{L}_{0,3}$};
			
			\node (plus) at (6.4, 8.1) {$\oplus$};
			\node (l) at (6.4, 8.85) {$\mathcal{L}$};
			
			\node[io] (o) at (10.8, 6.2) {output\\(112,1,166,144)};
			
			\foreach \from/\to in {x00/x10, x10/x20, x20/x30}
			\draw [->, brown] (\from) -- (\to);
			
			\foreach \from/\to in {x30/x21, x20/x11, x21/x12, x10/x01, x11/x02, x12/x03}
			\draw [->, teal] (\from) -- (\to);
			
			\foreach \from/\to in {x00/x01, x01/x02, x02/x03, x10/x11, x11/x12, x20/x21}
			\draw [->, blue] (\from) -- (\to);
			
			\foreach \from/\to in {x00/x02, x00/x03, x01/x03, x10/x12}
			\draw [->, blue] (\from) to[bend left] (\to);
			
			\draw [->] (i) -- (x00);
			\draw [->] (f03) -- (o);
			
			\draw [->, red] (x01) -- (f01);
			\draw [->, red] (x02) -- (f02);
			\draw [->] (x03) -- (f03);
			
			\draw [->, red] (f01) -- (l01);
			\draw [->, red] (f02) -- (l02);
			\draw [->, red] (f03) -- (l03);
			
			\draw [->, red] (l01) -- (plus);
			\draw [->, red] (l02) -- (plus);
			\draw [->, red] (l03) -- (plus);
			
			\draw [->, red] (plus) -- (l);
			
			\draw [->, red] (8.5, 1.75) -- (8.5, 2.15);
			\node[draw=none, fill=none, right] (l4) at (8.9, 1.95) {loss computation};
			\draw [->, blue] (8.3, 0) -- (8.7, 0);
			\node[draw=none, fill=none, right] (l3) at (8.9, 0) {skip connections};
			\draw [->, brown] (8.3, 0.85) -- (8.7, 0.45);
			\node[draw=none, fill=none, right] (l1) at (8.9, 0.65) {downsampling};
			\draw [->, teal] (8.3, 1.1) -- (8.7, 1.5);
			\node[draw=none, fill=none, right] (l2) at (8.9, 1.3) {upsampling};
		\end{tikzpicture}
		\subcaption{Detailed architecture overview: Each node $R$ is a residual block as depicted on the right.}
		\label{subfig:Overview}
	\end{subfigure}
	\hfill
	\begin{subfigure}[t]{0.32\textwidth}
		\centering
		\begin{tikzpicture}[scale=0.7, transform shape]
			\tikzstyle{convolution} = [rectangle, fill=orange!30, minimum width={110pt}]
			\tikzstyle{normalization} = [rectangle, fill=blue!30, minimum width={110pt}]
			\tikzstyle{activation} = [rectangle, fill=green!30, minimum width={110pt}]
			\tikzstyle{plain} = [draw=none, fill=none, align=center]
			
			\draw [fill=brown!10] (-2.5, -0.7) rectangle (2.7, 8.1);
			
			\node[plain] (input) at (0, 7.6) {input\\(B, M, 160, 144)};
			
			\node[convolution] (fusion) at (0, 6.4) {N $1 \times 1$ convolutions};
			\node[convolution] (conv1) at (0, 5.5) {N $3 \times 3$ convolutions};
			\node[normalization] (norm1) at (0, 4.9) {batch normalization};
			\node[activation] (act1) at (0, 4.3) {ReLu};
			\node[convolution] (conv2) at (0, 3.4) {N $3 \times 3$ convolutions};
			\node[normalization] (norm2) at (0, 2.8) {batch normalization};
			\node[plain] (plus) at (0, 1.9) {$\oplus$};
			\node[activation] (act2) at (0, 1.0) {ReLu};
			
			\node[plain] (output) at (0, -0.2) {output\\(B, N, 160, 144)};
			
			\draw (input) -- (fusion) node [midway, label={[font=\scriptsize, label distance=-0.4em]left:M}] {};
			\draw (fusion) -- (conv1) node [midway, label={[font=\scriptsize, label distance=-0.4em]left:N}] {};
			\draw (conv1) -- (norm1);
			\draw (norm1) -- (act1);
			\draw (act1) -- (conv2) node [midway, label={[font=\scriptsize, label distance=-0.4em]left:N}] {};
			\draw (conv2) -- (norm2);
			\draw (norm2) -- (plus) node [midway, label={[font=\scriptsize, label distance=-0.4em]left:N}] {};
			\draw (plus) -- (act2) node [midway, label={[font=\scriptsize, label distance=-0.4em]left:N}] {};
			\draw (act2) -- (output) node [midway, label={[font=\scriptsize, label distance=-0.4em]left:N}] {};
			
			\draw (0, 5.95) -- (2.4, 5.95) -- (2.4, 1.9) -- (plus) node [midway, label={[font=\scriptsize, label distance=-0.4em]below:N}] {};
		\end{tikzpicture}
		\subcaption{Architecture of a single residual block in our network.}
		\label{subfig:ResidualBlock}
	\end{subfigure}
	\caption[Architecture]{Architecture of Residual UNet++}
\end{figure}

The shape of our Residual UNet++ architecture is very similar to the original UNet++ architecture~\cite{zhou2018unet++}, but uses residual blocks including local skip connections as the basic building block of the network, as it was already proposed by Peng et al.~\cite{peng2019end}. The implementation is based on Python 3.7.3 using PyTorch 1.1.0.

Figure \ref{subfig:Overview} shows an overview of the complete network: Each node $R$ in this overview represents a single residual block. The input is a tensor of shape (B, C, W, H) with B denoting the batch size, C the number of feature channels, W and H the width and height of the image respectively. In our case, the input tensor has the shape (112, 10, 160, 144), and is fed to the first node $R_{0,0}$.
The final output of the network is a tensor of shape (B, 1, W, H) and will be retrieved from $F_{0,3}$.

Each diagonal arrow pointing downwards represents a downsampling step which is implemented using a maximum pooling operation with a $2 \times 2$ kernel and a stride of two.
This downsampling step divides the width and height of the image in halves, but at the same time doubles the width of the residual blocks.
The diagonal arrows pointing upwards represent the opposite operation, namely bilinear upsampling, doubling the width and height of the image.

Residual blocks at depth zero, i.e.~the nodes $R_{0,0}$ to $R_{0,3}$, have a width of 16 which gets doubled with increasing depth, leading to a width of 128 for $R_{3,0}$. If a residual block has two incoming edges, we simply concatenate the inputs before feeding them to the first layer of the residual block.
The final layers of the network, named $F_{0,1}$ to $F_{0,3}$, are simple convolutional layers with a $1 \times 1$ kernel returning a single feature map which is fed to a sigmoid function and compared to the desired target value. 

For the training phase, we use a form of deep supervision: We apply a sigmoid function to each final layer and sum up the losses $\mathcal{L}_{0,1}$ to $\mathcal{L}_{0,3}$ before backpropagation, indicated by the red arrows. For inference however, we only consider the output of the final layer $F_{0,3}$.

The architecture of a single residual block is depicted in Figure \ref{subfig:ResidualBlock}. 
The first convolution layer receives a tensor of shape (B, M, 160, 144) as input where M denotes the width of the tensor generated by concatenation of the previous layers' outputs. As the residual block is supposed to have a width of N, this first layer applies N $1 \times 1$ convolutions and outputs a tensor of shape (B, N, 160, 144) which is forwarded to the following layers, indicated by the small N next to each connection. The plus sign between the second batch normalization and the final activation simply sums the outputs of the second batch normalization and the first convolution layer.
All convolution layers apply zero padding to keep the original image dimensions.

If we consider $R_{0,1}$ as an example, we see that this node receives as input the outputs of $R_{0,0}$ with shape (B, 16, 160, 144) and $R_{1,0}$  with shape (B, 32, 160, 144) which are concatenated to a tensor of shape (B, 48, 160, 144) where 48 is the width M of the tensor in \autoref{subfig:ResidualBlock}. As the residual blocks of depth zero operate with a width of 16, the first layer of the residual block will apply 16 convolutions and generate a tensor of shape (B, 16, 160, 144). This step is necessary to avoid incompatible tensor shapes for the residual connection which adds the output of the first convolution layer to the output of the second batch normalization layer before the final activation.

\section{Experimental Setup}
\label{app:Training_Settings}

\begin{table}
	\centering
	\caption[Cross Validation Sets]{Cross validation sets and their number of samples.}
	\label{tab:Cross_Validation_Sets}
	\begin{tabular}{cccc}
		\toprule
		\multicolumn{3}{c}{Testing} & Training \\
		\cmidrule(r){1-3}
		Test Set & Time Range & \# Samples & \# Samples \\
		\midrule
		1 & 2017-06-01 00:30 to 2017-06-08 23:00 & 42668 & 53140 \\
		2 & 2017-06-09 11:00 to 2017-06-17 09:45 & 15960 & 79848 \\
		3 & 2017-06-17 21:45 to 2017-06-25 20:15 & 12096 & 85000 \\
		4 & 2017-06-26 08:15 to 2017-07-04 06:30 & 23740 & 73356 \\
		\bottomrule
	\end{tabular}
\end{table}

All our experiments have been conducted on a computer equipped with two Intel Xeon E5-2600 v4 processors with a total of 32 GB of RAM. To accelerate the training, we used two Nvidia GTX 1080 Ti graphics cards.

We have fixed the number of epochs to 30, resulting in a training time of roughly 17 hours per cross validation step. The learning rate was initialized with a value of 0.01 and decreased by a factor of 10 if the training loss showed no relative improvement of 1\% during the last 5 epochs. The optimizer chosen was Stochastic Gradient Decent with a weight decay factor of 0.1. The batch size has been fixed to 112, i.e.~the number of samples generated from two original images.

The loss function $\mathcal{L}$ used during training was weighted binary cross entropy. It is defined as a final sigmoid activation followed by a binary cross entropy computation where a special weight is applied to the positive class:

\begin{equation*}
	\mathcal{L}_{n} = -(p \cdot y_n \cdot log(\sigma(x_n)) + (1 - y_n) \cdot log(1 - \sigma(x_n)))
\end{equation*}

In the above formula, $n$ denotes the number of the current sample and $\sigma$ the sigmoid activation function applied to the model outputs $x_n$. $y_n$ is the true class of the sample and $p$ the weight applied to the positive class.

The cross validation sets are built by diving the data into four time ranges of equal size, depicted in Table \ref{tab:Cross_Validation_Sets}. Each test set covers the data of roughly eight days. The corresponding training sets have been built by taking all remaining data minus a twelve hour margin around each test set to avoid unwanted cross correlation effects between training and test set. For the first test set, this means that the model was trained using all data from 2017-06-09 11:00 to 2017-07-04 06:30. The resulting number of samples in each training and test set is also given in the table. The imbalance in size between the different sets is due to corrupt satellite image files.

\section{Evaluation Metrics}
\label{app:Evaluation_Metrics}

\begin{table}
	\centering
	\caption[Confusion Matrix]{Possible outcomes of the confusion matrix.}
	\label{tab:Confusion_Matrix}
	\begin{tabular}{c|c|c}
		\toprule
		 & \multicolumn{2}{c}{True class} \\
		Predicted class & Positive & Negative \\
		\midrule
		Positive & True Positive (TP) & False Positive (FP) \\
		Negative & False Negative (FN) & True Negative (TN) \\
		\bottomrule
	\end{tabular}
\end{table}

To evaluate our approach, we use four different evaluation metrics, namely True Positive Rate, True Negative Rate, Accuracy and False Alarm Ratio.
All of them can be computed by considering the values of the confusion matrix.
To generate this matrix, we compare the predicted class of each pixel on each image with the true class it is supposed to have, i.e.~we perform a strict matching between prediction and true class. 
Depending on the result, we distinguish four outcomes of this comparison which are denoted in \autoref{tab:Confusion_Matrix}: True Positives (TP), False Positives (FP), False Negatives (FN) and True Negatives (TN).
Each cell of the confusion matrix finally contains the number of samples fulfilling the corresponding combination of predicted and true class.

The four evaluation metrics used in our paper are then defined as follows:

\begin{itemize}
	\item True Positive Rate (also called Probability of Detection, POD): $\frac{\text{TP}}{\text{TP + FN}}$
	\item True Negative Rate: $\frac{\text{TN}}{\text{TN + FP}}$
	\item Accuracy: $\frac{\text{TP + TN}}{\text{TP + TN + FP + FN}}$
	\item False Alarm Ratio: $\frac{\text{TP}}{\text{TP + FP}}$
\end{itemize}

\end{document}